\documentclass[10pt,conference]{IEEEtran}
\IEEEoverridecommandlockouts
\usepackage[T1]{fontenc}
\usepackage{cite}
\usepackage{amsmath,amssymb,amsfonts}
\usepackage{algorithmic}
\usepackage{graphicx}
\usepackage{textcomp}
\usepackage{verbatim}
\usepackage[utf8]{inputenc}
\usepackage{pmboxdraw}
\usepackage{float}
\usepackage{tikz}
\usepackage{xcolor}
\usepackage{mdframed}
\usepackage[hyphens]{url}
\usepackage{hyperref}
\usepackage{multirow}
\usepackage{fontawesome}

\newcommand{\IEEEpreprintcopyright}{%
\IEEEpubid{%
\raisebox{-11mm}{
\makebox[\textwidth]{\hfill%
\fcolorbox{red}{white}{%
\parbox{0.8\textwidth}{\footnotesize
\textcopyright\ \the\year\ IEEE. Personal use of this material is permitted.
Permission from IEEE must be obtained for all other uses, including
reprinting/republishing this material for advertising or promotional purposes,
collecting new collective works for resale or redistribution to servers or lists,
or reuse of any copyrighted component of this work in other works.
}%
}%
\hfill}%
}
}%
}

\def\BibTeX{{\rm B\kern-.05em{\sc i\kern-.025em b}\kern-.08em
    T\kern-.1667em\lower.7ex\hbox{E}\kern-.125emX}}
\begin{document}

\title{%
{\normalfont\footnotesize\itshape
Accepted for publication at the 2026 IEEE World Congress on Computational Intelligence (WCCI 2026)\par}
\vspace{0.5em}
A Two-Stage LLM Framework for Accessible and Verified XAI Explanations\\
\thanks{This work has been partially supported by project MIS 5154714 of the National Recovery and Resilience Plan Greece 2.0, funded by the European Union under the NextGenerationEU Program. The first two authors contributed equally to this work.}
\thanks{The authors gratefully acknowledge the Computer Centre of the Department of Computer Engineering \& Informatics at the University of Patras for providing the computational resources that supported this research.}
}

\author{
    \IEEEauthorblockN{Georgios Mermigkis\textsuperscript{†,¶}, 
    Dimitris Metaxakis\textsuperscript{†,¶}, 
    Marios Tyrovolas\textsuperscript{‡,§,¶}, 
    Argiris Sofotasios\textsuperscript{†,¶},
    Nikolaos Avgeris\textsuperscript{‡,¶}, \\
    Panagiotis Hadjidoukas\textsuperscript{†,§},
    Chrysostomos Stylios\textsuperscript{‡,§}}    
    \vspace{0.5em}
    
    \IEEEauthorblockA{\textsuperscript{†}Department of Computer Engineering and Informatics, University of Patras, Patras, Greece}
    \IEEEauthorblockA{\{g.mermigkis, d.metaxakis, a.sofotasios, phadjido\}@ceid.upatras.gr}
    
    \IEEEauthorblockA{\textsuperscript{‡}Department of Informatics and Telecommunications, University of Ioannina, Arta, Greece}
    \IEEEauthorblockA{\{tirovolas, avgeris.nik\}@kic.uoi.gr}

    \IEEEauthorblockA{\textsuperscript{§}Industrial Systems Institute, Athena Research Center, Patras, Greece}
    \IEEEauthorblockA{stylios@athenarc.gr}
    
    \IEEEauthorblockA{\textsuperscript{¶}Archimedes Unit, Athena Research Center, Athens, Greece}

}

\maketitle
\IEEEpreprintcopyright

\begin{abstract}
Large Language Models (LLMs) are increasingly used to translate the technical outputs of eXplainable Artificial Intelligence (XAI) methods into accessible natural-language explanations. However, existing approaches often lack guarantees of accuracy, faithfulness, and completeness. At the same time, current efforts to evaluate such narratives remain largely subjective or confined to post-hoc scoring, offering no safeguards to prevent flawed explanations from reaching end-users. To address these limitations, this paper proposes a Two-Stage LLM Meta-Verification Framework that consists of (i) an Explainer LLM that converts raw XAI outputs into natural-language narratives, (ii) a Verifier LLM that assesses them in terms of faithfulness, coherence, completeness, and hallucination risk, and (iii) an iterative refeed mechanism that uses the Verifier’s feedback to refine and improve them. Experiments across five XAI techniques and datasets, using three families of open-weight LLMs, show that verification is crucial for filtering unreliable explanations while improving linguistic accessibility compared with raw XAI outputs. In addition, the analysis of the Entropy Production Rate (EPR) during the refinement process indicates that the Verifier’s feedback progressively guides the Explainer toward more stable and coherent reasoning. Overall, the proposed framework provides an efficient pathway toward more trustworthy and democratized XAI systems.
\end{abstract}


\section{Introduction}

The field of eXplainable Artificial Intelligence (XAI) has advanced rapidly in recent years, delivering a diverse set of techniques that clarify how and why AI models reach their decisions \cite{Mersha2024-ra}. Current methods usually highlight the most influential input features \cite{Lundberg2017}, whereas others pinpoint image regions that guide predictions in computer vision tasks \cite{Selvaraju2017-fz} or generate counterfactual explanations that describe how minimal input changes would alter model outputs \cite{Tyrovolas2026-hi, Wachter2018}. While these approaches enhance algorithmic transparency, their outputs are often too technical for domain experts, decision-makers and general end-users, limiting their accessibility. Large Language Models (LLMs) offer a promising tool for bridging this gap. Their ability to perform contextual reasoning, summarize complex information and generate fluent natural-language text positions them as effective mediators capable of translating technical XAI outputs into coherent and human-readable narratives \cite{Zytek2024-yz}.

Several recent studies have begun exploring this novel paradigm \cite{Silvestri2025-hz}. For example, the authors in \cite{Mavrepis2024-po} introduced x-[plAIn], a customized GPT-based framework that interprets outputs from different XAI techniques, including LIME, SHAP, Grad-CAM, and PDP, targeting both technical and non-technical audiences. Castellano \textit{et al.} \cite{castellano2024using} applied LLMs to explain Grad-CAM heatmaps in AI-generated art classification, illustrating how visual saliency maps can be transformed into clearer justifications. Similarly, Zeng \textit{et al.} \cite{Zeng2024-ep} used the Mistral-7B model to convert SHAP feature attribution values into accessible natural-language narratives. However, current architectures treat explanation generation as a single-step process and lack mechanisms for systematically validating the accuracy and completeness of the produced narratives. This is critical because recent theoretical analyses have shown that hallucinations, omissions, and reasoning failures are inherent limitations of LLMs, potentially leading to misleading or overly persuasive explanations \cite{Kalai2025-ok}.

In response, a growing body of research has investigated methods for evaluating LLM-generated XAI narratives. Early efforts relied on human judgments of faithfulness, whereas newer approaches employ LLM-based evaluators or automated frameworks that compare explanations against expert-written references or raw XAI outputs \cite{Castelnovo2024-im, Shirvani-Mahdavi2025-ze, Ichmoukhamedov2024-vs}. However, human evaluations are often influenced by subjective expectations, while expert-written gold standards may reflect stylistic conventions that blur the distinction between narrative quality and true explanatory accuracy \cite{Silvestri2025-hz}. Although LLM-based evaluators provide a middle ground, they have primarily been used for scoring and benchmarking, rather than as a general, method-agnostic verification-and-repair stage that systematically detects and corrects errors before delivery. Without such safeguards, flawed narratives may still reach end-users.

To address these limitations, this paper introduces a Two-Stage LLM Meta-Verification Framework for generating accessible and reliable natural-language explanations from XAI methods. The framework consists of an explanation generation stage, followed by a verification stage that systematically evaluates the faithfulness, logical coherence, completeness and hallucination risk of produced narratives before they are delivered to the user. In addition, unlike previous studies, it incorporates an iterative refeed mechanism that leverages verification feedback to revise and refine explanations. The approach is empirically validated across multiple XAI techniques, datasets and open-weight LLM families, under both natural and synthetic error conditions. The results demonstrate substantial improvements in explanation reliability, contributing to the advancement of more transparent, trustworthy and broadly accessible XAI systems.

\section{The Two-Stage LLM Meta-Verification Framework} \label{proposed_framework}

\begin{figure}[htbp!]
    \centering
    \includegraphics[width=\columnwidth]{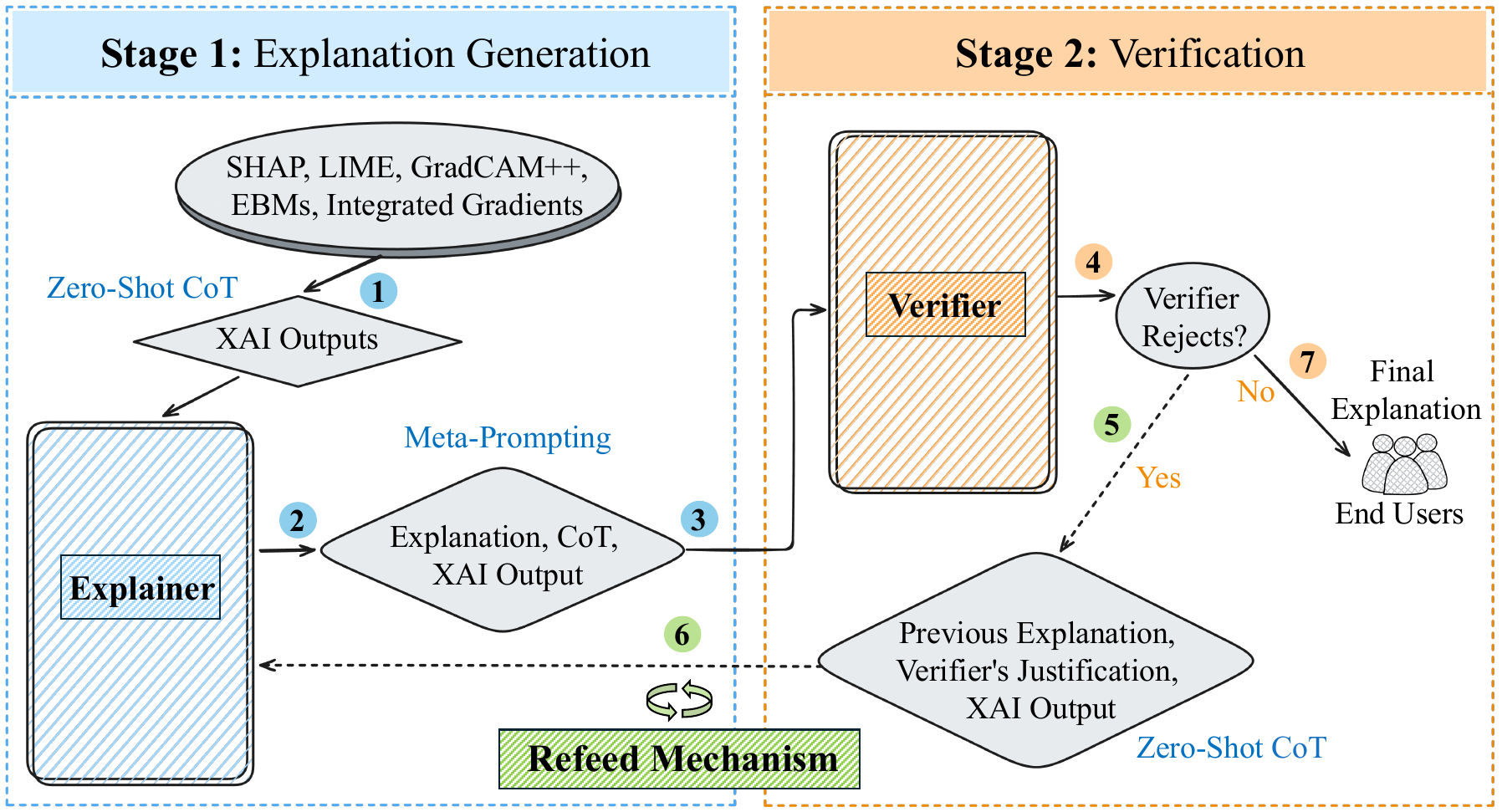}
    \caption{Overview of the proposed Two-Stage LLM Meta-Verification framework. Stage~1 (Explainer) generates a natural-language explanation from XAI outputs, while Stage~2 (Verifier) evaluates the explanation and issues an accept--reject verdict; rejected explanations are refined via the refeed mechanism.}
    \label{fig:framework}
\end{figure}

This section presents the proposed Two-Stage LLM Meta-Verification Framework. As illustrated in Fig.~\ref{fig:framework}, the process consists of two stages: (1) an Explainer LLM that generates a natural-language explanation from XAI outputs, and (2) a Verifier LLM that evaluates the explanation, identifies errors, and decides whether it should be accepted or revised. The following subsections describe these components and their interaction.

\subsection{Explainer Component}

The Explainer is an LLM that transforms XAI outputs (e.g., feature importance scores or saliency maps) into accurate, human-readable explanations. Its behavior is guided by a prompt template tailored to the specific underlying XAI method. This template defines the Explainer’s role and expertise, provides method-specific guidelines, and incorporates a refusal block that prevents instruction leakage. It also supplies the relevant dataset context and a brief description of the XAI technique and concludes by explicitly stating the explanation task and supplying the XAI output for interpretation.

To enhance reasoning quality, the Explainer employs a Zero-Shot Chain-of-Thought (CoT) prompting strategy \cite{Kojima2022-mw}, which asks the model to generate intermediate reasoning steps before producing the final explanation. In Zero-Shot CoT, the model is not shown any task-specific examples; instead, a generic cue such as “Let’s think step by step” is embedded in the prompt, triggering the model to generate a structured reasoning process, as outlined in the left panel of Fig.~\ref{fig:explainer-verifier-templates}. This simple yet effective technique has been shown to substantially improve performance on multi-step reasoning tasks, including arithmetic, symbolic, and logical problems \cite{Kojima2022-mw}.

\subsection{Verifier Component}

The Verifier LLM evaluates the explanations produced by the Explainer and determines whether they should be delivered to end-users or revised. It is driven by a structured meta-prompting template that encodes reusable high-level instructions specifying which aspects of an explanation must be assessed, in what order, and how the final judgment should be formatted. Unlike few-shot prompting, which relies on imitating specific input-output examples and often lacks robustness across tasks \cite{Brown2020}, the meta-prompt formalizes the evaluation procedure itself \cite{Zhang2023-pu}, enabling generalization across heterogeneous XAI methods without method-specific templates.  

\begin{figure*}[t!]
    \centering
\includegraphics[width=1\textwidth]{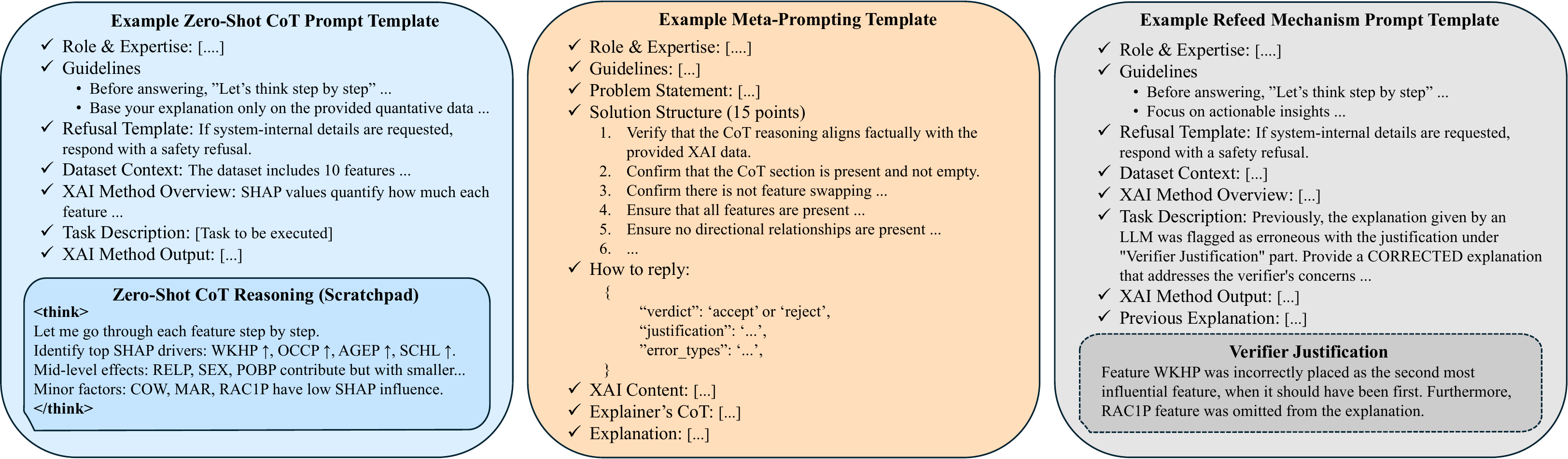}
    \caption{Structured prompt templates used in the framework: Explainer Zero-Shot CoT (left), Verifier Meta-Prompting (center) and Refeed Mechanism for iterative correction (right).}
    \label{fig:explainer-verifier-templates}
\end{figure*}

In this setting, the Verifier is assigned the role of an impartial critic and is required to reason explicitly through four core criteria: the faithfulness of the explanation to the underlying attributions, its internal logical coherence, its completeness with respect to the XAI method's output, and the absence of hallucinations or unsupported causal mechanisms. Only after completing this structured reasoning phase is the Verifier permitted to issue a final verdict. To support automation and systematic analysis, the Verifier returns a standardized response that contains a binary accept-reject decision, a concise justification grounded in the reasoning process, and an error type selected from a set of six predefined categories described in Section~\ref{error-spaces}. The structure of the meta-prompting template guiding this procedure is shown in the central panel of Fig.~\ref{fig:explainer-verifier-templates}.

\subsection{Refeed Mechanism}

When the Verifier rejects an explanation, it returns targeted feedback that is injected back into the Explainer’s prompt through a refeed mechanism. The revised prompt preserves the structure of the initial template and still relies on Zero-Shot CoT reasoning, but its task description is updated to explicitly reference the identified error and request a corrected explanation. The prompt is completed with dynamically populated fields—placeholders that are automatically filled with the previous explanation and the Verifier’s justification—so that each new attempt directly addresses the specific error detected in the previous pass, as illustrated in the right panel of Fig.~\ref{fig:explainer-verifier-templates}. This process continues until the explanation is accepted, or the maximum number of trials, $K$, is reached.

\section{Experimental Setup} \label{experimental_setup}


This section describes the datasets, XAI methods, and Machine Learning (ML) models, along with the LLM pairs, error types, and evaluation metrics used in the proposed framework. All experiments were conducted in Python 3.10 on a system with a dual-socket Intel Xeon Silver 4116 CPU (24 cores, 48 threads), an NVIDIA Tesla V100 GPU (32 GB VRAM), and 128 GB RAM. The source code is available at \href{https://github.com/gMerm/Two-Stage-LLM-Framework-For-Verified-XAI-Explanations}{\faGithub}.

\begin{figure*}[htbp!]
    \centering
\includegraphics[width=\textwidth]{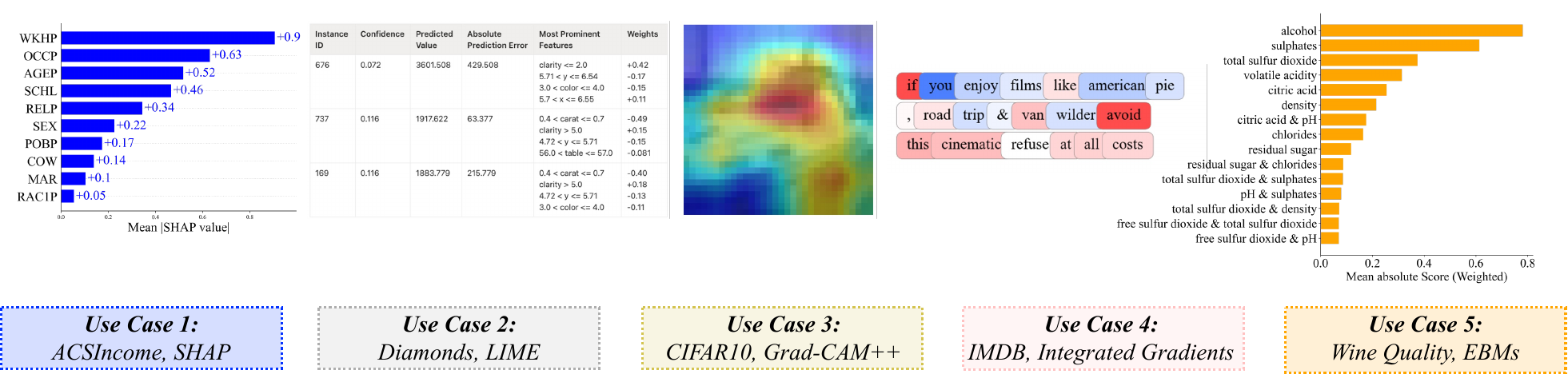}
    \caption{Representative explanations from the five use cases: SHAP on ACSIncome, LIME on Diamonds, Grad-CAM++ on CIFAR10, Integrated Gradients on IMDB Reviews and Explainable Boosting Machines on Wine Quality.}
    \label{fig:xai-examples}
\end{figure*}

\subsection{Use Cases: XAI Methods and Datasets}

To cover a broad spectrum of XAI methods, we employed five well-established datasets (Fig. ~\ref{fig:xai-examples}) spanning diverse data modalities, including tabular, image, and textual data. Each dataset was paired with a specific ML model and explanation technique, ranging from feature attribution vectors to spatial saliency maps. Despite this heterogeneity, all explanation outputs were converted into text before processing. Moreover, the experimental setup included both post-hoc explanation methods and intrinsically interpretable models. Specifically, the following five use cases were investigated:

\textbf{ACSIncome.}  
The ACSIncome dataset comprises 1.6 million samples with 10 predictive features and a continuous income target (i.e., PINCP). Following prior recommendations on replacing the problematic UCI Adult dataset \cite{Ding2021}, the task was converted into binary classification by thresholding PINCP at its median value. Subsequently, an XGBoost classifier was trained and SHAP was applied to derive global feature importance scores.

\textbf{CIFAR-10.}
The CIFAR-10 dataset consists of 60,000 color images of size $32\times32$ across 10 object classes. For image classification, we employed a ResNet-20 convolutional neural network pre-trained on CIFAR-10 and loaded it using the PyTorch Hub. Subsequently, to explain the model’s predictions, we generated Grad-CAM++ heatmaps that identified the most influential image regions for each decision. These visual explanations were then converted into deterministic, structured textual descriptions by extracting activation statistics and coarse spatial localization information using a fixed low-level vocabulary (e.g., \emph{top--left}, \emph{center}, \emph{bottom--right}). The resulting text-based descriptions were subsequently passed to the Explainer--Verifier pipeline.

\textbf{IMDB Reviews.}  
The IMDB dataset contains 50,000 movie reviews labeled as positive or negative. An LSTM-based sentiment classifier was trained on this dataset, and the Integrated Gradients (IG) method was used to compute token-level attributions by interpolating from a zero-embedding baseline to the true input. These explanations are provided to the Explainer--Verifier pipeline.

\textbf{Diamonds.}  
The Diamonds dataset comprises approximately 54,000 samples with nine predictive features and a continuous price target. An XGBoost regressor was trained, and LIME was subsequently applied to generate local explanations. These local attributions were aggregated across multiple validation instances to derive approximate global feature importance scores. The resulting explanations were then used as input to the Explainer--Verifier pipeline.

\textbf{Wine Quality.}  
The Wine Quality dataset consists of 12 physicochemical features describing the properties of red and white wines. The target quality score was binarized into high-quality ($\ge 7$) and low-quality classes, and an Explainable Boosting Machine (EBM) classifier was trained. The EBM inherently provides global feature importance scores, capturing the primary factors driving the model’s predictions.

\subsection{LLM Selection and Inference Settings}
Three open-weight model families were selected as the underlying LLMs for the proposed framework: \textit{GPT-OSS}, \textit{DeepSeek-R1}, and \textit{Qwen-3}. From each family, we chose models at the lower end of the deployment spectrum (14B–30B parameters), striking a practical balance between reasoning capability and computational efficiency, thus making them suitable for resource-constrained on-premises or cloud environments. Moreover, the selected models reflect distinct training paradigms. \textit{GPT-OSS} emphasizes openness and balanced general-purpose reasoning \cite{OpenAI2025-tw}, \textit{DeepSeek-R1} incorporates reinforcement-style training signals to strengthen stepwise reasoning \cite{DeepSeek-AI2025-vx}, while \textit{Qwen-3} is optimized for structured instruction following and demonstrates strong multilingual capabilities \cite{Yang2025-ne}. 

All models were executed in inference-only mode using quantized weights via the Ollama runtime. To ensure a fair and consistent comparison across all evaluated models, a unified inference configuration was adopted. Generation parameters were fixed to a temperature of 0.6, maximum generation length of 2{,}048 tokens, and context window of 128K tokens. In addition, all models were executed with the reasoning mode (“think”) enabled.

\subsection{Error Spaces}
\label{error-spaces}
Since LLM-generated narratives can fail in diverse and unpredictable ways, the framework must be tested across varying levels of difficulty, frequency, and structural complexity to ensure robust and generalizable evaluation. To this end, we defined two complementary error spaces capturing both realistic and systematic failure modes.

\textbf{Natural Error Space.} This space captures naturally occurring errors arising during explanation generation under standard operating conditions, including incomplete attributions, unsupported causal claims, factual drifts, and structural omissions. It was constructed by executing the Explainer--Verifier pipeline in a controlled loop until either 1,000 verified instances were collected or 200 outputs were rejected. To ensure balanced representation and avoid bias toward specific methods or datasets, a round-robin scheduling strategy was applied across the five XAI methods and datasets.

\textbf{Synthetic Error Space.} To ensure full coverage of all predefined error categories, including rare ones, a mutation-based synthetic space was constructed. Starting from valid explanations, mutation operators perturb them along six orthogonal axes: SwapTopFeature, SwapMinorFeature, NegateRelation, OmitFeature, InsertHallucination, and TruncateResponse, each representing a distinct failure mode.

\subsection{Evaluation Metrics}
The framework’s performance was primarily evaluated based on its ability to detect erroneous explanations. Detection accuracy was computed using true positives (\textit{TP}: faithful explanations correctly accepted), true negatives (\textit{TN}: erroneous explanations correctly rejected), false positives (\textit{FP}: faithful explanations incorrectly rejected), and false negatives (\textit{FN}: erroneous explanations incorrectly accepted), which capture how reliably the Verifier distinguishes valid from invalid explanations.

To assess linguistic accessibility, the Flesch-Kincaid metrics, namely Reading Ease and Grade Level, were used \cite{kincaid1975readability}. Finally, uncertainty was measured using the Entropy Production Rate (EPR), derived from the entropy dynamics of the top-10 token-level log probabilities generated by the LLMs. Lower values indicate more stable outputs, whereas higher values reflect increased uncertainty \cite{Moslonka2025-ks}.

\begin{figure*}[t!]
    \centering
\includegraphics[width=\textwidth]{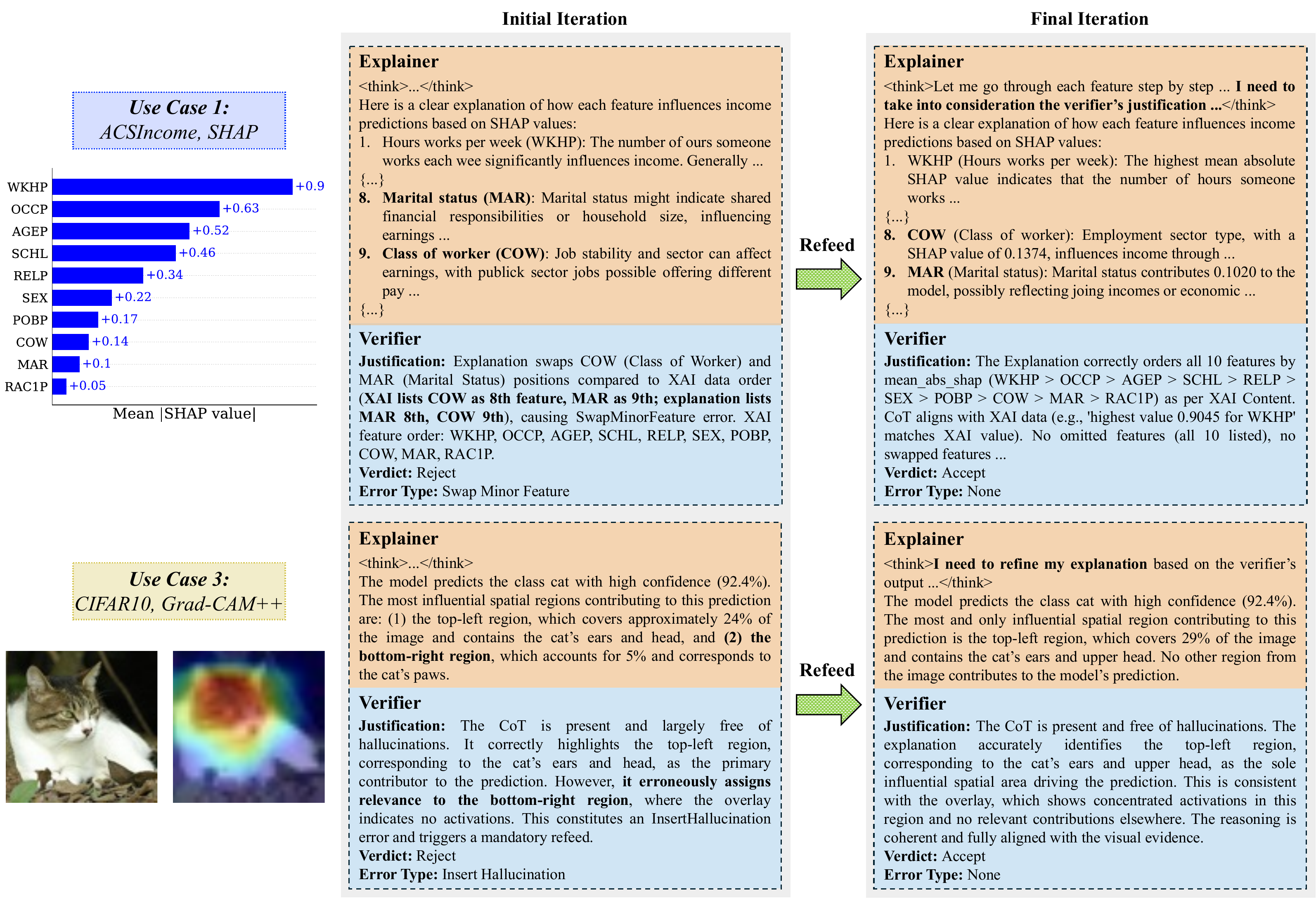}
    \caption{Illustrative examples of the proposed framework correcting erroneous explanations after verifier-guided refinement.}
    \label{fig:qualitative-examples}
\end{figure*}

\section{Results and Discussion}

This section presents and analyzes the empirical results of this study. First, to mitigate shared hallucinations, correlated errors, and self-preference biases that arise when a single LLM performs both roles \cite{Panickssery2024}, we employed different models for the Explainer and Verifier. Fig.~\ref{fig:qualitative-examples} illustrates representative examples for use cases 1 and 3 after applying the framework, where initial explanations exhibited failures such as SwapMinorFeature and InsertHallucination, which were corrected following verification.

To evaluate the proposed method, we addressed the following four research questions (RQs):
\begin{enumerate}
    \item RQ1: Does the architecture reduce the number of erroneous explanations that would otherwise reach the end user?
    \item RQ2: Does structured meta-prompting improve the reliability of the Verifier in detecting erroneous explanations?
    \item RQ3: Does the framework improve the linguistic accessibility of the explanations?
    \item RQ4: Do uncertainty signals, captured through EPR, reflect convergence and explanation stability?
\end{enumerate}

Concerning RQ1, Table~\ref{tab:explainer-verifier-results} reports the performance of all examined Explainer--Verifier pairs in the Natural Error Space experiment, as this setting is more representative of real-world use than synthetic perturbations. In this experiment, each Explainer generated outputs until either 1,000 verified explanations or 200 were rejected by an initial Verifier. In practice, all the Explainers reached the rejection threshold before attaining the acceptance target, making the 200-error limit the actual stopping criterion. The collected explanations were manually annotated by the authors according to a predefined protocol, with any disagreements resolved through discussion until a consensus was achieved. Finally, the annotated explanations were reassessed using two alternative Verifiers to evaluate the classification accuracy and robustness across the verifier configurations.

\begin{table*}[t!]
\centering
\caption{Performance of cross-model Explainer--Verifier LLM pairs in the Natural Error Space experiment.}
\label{tab:explainer-verifier-results}

\resizebox{0.95\textwidth}{!}{%
\renewcommand{\arraystretch}{1.3}
\begin{tabular}{l l c c c c c c c c c}
\hline
\textbf{Explainer} & \textbf{Verifier} & \textbf{\#Samples} 
& \textbf{TP} & \textbf{TN} & \textbf{FP} & \textbf{FN} 
& \textbf{Explainer} & \textbf{Explainer} 
& \textbf{Verifier} & \textbf{Verifier} \\
 &  &  
 &  &  &  &  
 & \textbf{Err.\%} & \textbf{Only Acc.\%} 
 & \textbf{Acc.\%} & \textbf{F1\%} \\
\hline
deepseek-r1:14b & gpt-oss:20b & 720 & 467 & 181 & 53 & 19 & 32.5 & 67.5 & 90.0 & 92.86 \\
deepseek-r1:14b & qwen3:30b  & 720 & 406 & 183 & 51 & 80 & --//-- & –//– & 81.8 & 86.08 \\
\underline{gpt-oss:20b} & \textbf{qwen3:30b}  & 940 & 713 & 182 & 27 & 18 & \underline{22.23} & \underline{77.8} & \textbf{95.21} & \textbf{96.94} \\
gpt-oss:20b & deepseek-r1:14b & 940 & 708 & 9 & 200 & 23 & --//-- & –//– & 76.27 & 86.4 \\
qwen3:30b & gpt-oss:20b & 459 & 241 & 170 & 18 & 30 & 40.95 & 59 & 89.54 & 90.94 \\
qwen3:30b & deepseek-r1:14b & 459 & 268 & 4 & 184 & 3 & --//-- & –//– & 59.25 & 74.1 \\
\hline
\end{tabular}
}

\vspace{0.4em}
\makebox[0.95\textwidth][l]{%
    \hspace{0.4em}%
    \footnotesize\textit{Notes:} Bold entries indicate the best Verifier; underlined entries indicate the best Explainer.%
}
\makebox[0.95\textwidth][l]{%
    \hspace{3.35em}%
    \footnotesize The relationship between columns is $\#\text{Samples} = \text{TP} + \text{TN} + \text{FP} + \text{FN}$.%
}
\end{table*}

The results revealed clear performance differences among the three model families. As Explainers, the models varied substantially: \texttt{gpt-oss:20b} produced the lowest error rate (22.23\%; 209 failures out of 940 outputs), followed by \texttt{deepseek-r1:14b} (32.5\%), while \texttt{qwen3:30b} performed the worst at 40.95\%. However, when acting as Verifiers, the ranking is reversed. Specifically, \texttt{qwen3:30b} was the most reliable critic, achieving 95.21\% accuracy and 96.94\% F1-score when evaluating explanations from \texttt{gpt-oss:20b}. The second-best Verifier is \texttt{gpt-oss:20b}, which achieves 90\% accuracy and a 92.86\% F1-score when verifying \texttt{deepseek-r1:14b}. In contrast, \texttt{deepseek-r1:14b} performs poorly as a Verifier, tending to over-accept incorrect explanations and offering limited safety guarantees; therefore, it is excluded from subsequent Verifier analyses. Overall, these findings identify \texttt{gpt-oss:20b} as the most reliable Explainer and \texttt{qwen3:30b} as the strongest Verifier and their combination yields the highest end-to-end verification accuracy (95.21\%).

The “Explainer Only Acc.\%” results further highlight the necessity of verification: without it, Explainer accuracy ranges only from 59\% to 77.8\%. For the best-performing Explainer--Verifier pair, meta-verification increases accuracy from 77.8\% to 95.21\%, blocking 164 out of 209 erroneous explanations (a 78.5 \% reduction) from reaching the end user. This demonstrates that verification is not merely beneficial but essential for producing trustworthy and faithful explanations, particularly when the Explainer exhibits a non-negligible baseline error rate. These findings are also consistent with architectural differences across families: \textit{GPT-OSS}'s design for broad general-purpose reasoning supports strong explanatory reliability \cite{OpenAI2025-tw}; \textit{DeepSeek-R1}’s forward chain-of-thought enables moderate but inconsistent explanatory quality \cite{DeepSeek-AI2025-vx}; and \textit{Qwen-3}'s hybrid Mixture of Experts (MoE) and unified thinking design equips it with the evaluative stability needed for high-precision verification \cite{Yang2025-ne}.

Subsequently, to investigate the effect of structured meta-prompting on verification performance (RQ2), we conducted an ablation study on the Synthetic Error Space in which the refeed mechanism was disabled, enforcing a single-pass evaluation. Under this setting, all examined Verifier models were evaluated using three progressively simplified prompt configurations. The first configuration, \textbf{V0}, corresponds to the full meta-prompting template (Fig.~\ref{fig:explainer-verifier-templates}). The second, \textbf{V1}, removes the 15 core meta-prompting instructions while retaining the rest of the prompt structure. The third, \textbf{V2}, further reduces the prompt to a minimal version containing only essential task information, namely the Verifier’s role and expertise, the problem statement, and the expected response format.

The results (Table~\ref{tab:ablation-verifiers}) show that \textbf{V0} achieves the highest performance in terms of both accuracy and F1-score, while the performance degrades under \textbf{V1} and remains similarly reduced under \textbf{V2}. Among the evaluated models, \texttt{qwen3:30b} demonstrated the best performance, indicating its robustness to prompt simplification. In contrast, \texttt{deepseek-r1:14b} performed the worst across all configurations, highlighting both its sensitivity to prompt design and limited effectiveness as a Verifier. Overall, these findings indicate that removing the reasoning guidance provided by meta-prompting negatively impacts the Verifier’s ability to reliably assess explanations.

\begin{table}[htbp!]
\centering
\caption{Ablation results for Verifier models across prompt configurations (V0–V2) under single-pass evaluation on the Synthetic Error Space.}
\label{tab:ablation-verifiers}

\resizebox{\columnwidth}{!}{%
\renewcommand{\arraystretch}{1.3}
\begin{tabular}{l c c c c c c}
\hline
\multirow{2}{*}{\textbf{Verifier}} & \multicolumn{3}{c}{\textbf{Accuracy (\%)}} & \multicolumn{3}{c}{\textbf{F1-score (\%)}} \\
\cline{2-7}
 & \textbf{V0} & \textbf{V1} & \textbf{V2} & \textbf{V0} & \textbf{V1} & \textbf{V2} \\
\hline
qwen3:30b & \textbf{100} & 93.3 & 93.3 & \textbf{99.9} & 96.5 & 96.5 \\
gpt-oss:20b & \textbf{96.7} & 90.0 & 90.0 & \textbf{98.3} & 94.7 & 94.7 \\
deepseek-r1:14b & \textbf{40.0} & 23.3 & 30.0 & \textbf{57.1} & 37.8 & 46.15 \\
\hline
\end{tabular}
}
\end{table}

To assess readability and thus accessibility (RQ3), Flesch–Kincaid metrics were computed for verified explanations. Because the Verifier acts solely as a filtering mechanism and does not modify the explanations’ wording, these scores directly capture the linguistic quality produced by each Explainer. According to the results, raw XAI outputs are highly technical, with an average Reading Ease of 18.53 and Grade Level of 21.79, implying the need for postgraduate reading ability. In contrast, all Explainers significantly improved these scores. Among them, \texttt{gpt-oss:20b} yielded the most accessible explanations, achieving an average Reading Ease of 34.93 and an average Grade Level of 12.94, corresponding to an improvement of +16.40 Reading Ease points and a reduction of 8.85 grade levels relative to the baseline. Table~\ref{tab:readability-results} summarizes the readability improvements across all models. Therefore, the framework ensures that the delivered explanations are not only faithful to the underlying XAI outputs but also clearer and more accessible to end-users.

\begin{table}[htbp!]
\centering
\caption{Average Readability scores of Explainer LLMs and their improvements relative to raw XAI outputs.}
\label{tab:readability-results}

\resizebox{\columnwidth}{!}{%
\renewcommand{\arraystretch}{1.3}
\begin{tabular}{lcc}
\hline
\textbf{Explainer} &
\textbf{Flesch Reading Ease} ($\uparrow$) &
\textbf{Flesch--Kincaid Grade} ($\downarrow$) \\
\hline
Raw XAI Outputs
    & 18.53 
    & 21.79 \\
gpt-oss:20b       
    & \textbf{34.93} (+16.40) 
    & \textbf{12.94} (-8.85) \\
qwen3:30b         
    & 28.93 (+10.40) 
    & 13.00 (-8.79) \\
deepseek-r1:14b   
    & 23.35 (+4.82)  
    & 14.54 (-7.25) \\
\hline
\end{tabular}
}

\end{table}

The second part of the evaluation assessed the effectiveness of the refeed mechanism on 65 erroneous cases (35 randomly sampled from the Natural Error Space and 30 synthetic cases). This number of samples is statistically sufficient to support reliable estimation, yielding an average convergence rate of 96.92\% (63/65). The corresponding Agresti--Coull 95\% confidence interval is [88.7\%, 99.7\%], with an approximate error margin of 5.51\%. Each case was allowed up to 10 refinement iterations.

\begin{figure}[t!]
    \centering
    \begin{minipage}{1\columnwidth}
        \centering
        \includegraphics[width=\linewidth]{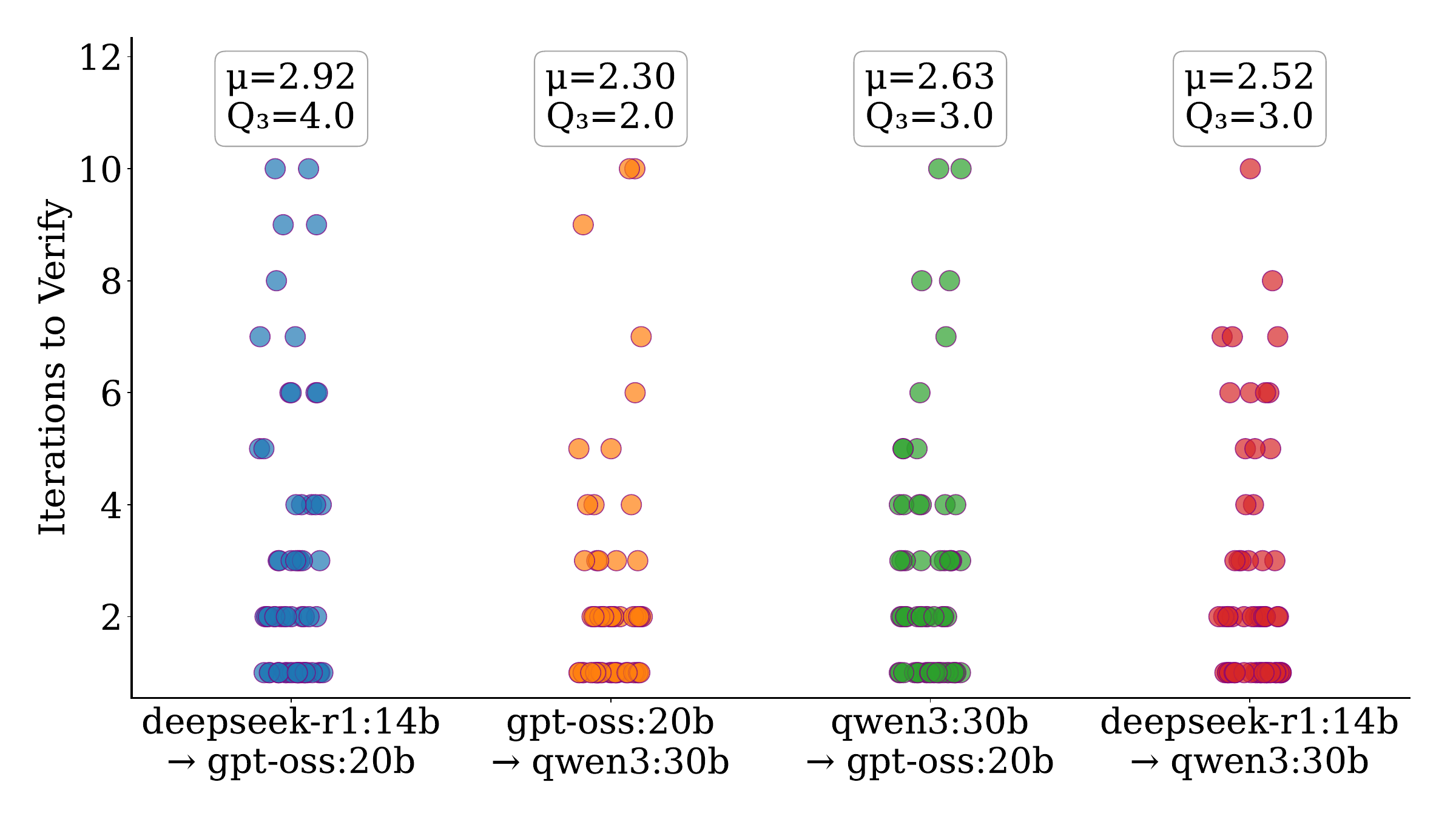}
    \end{minipage}
    
    \caption{Iterations required for explanation acceptance across the four Explainer--Verifier pairs. Each point represents one of the 65 cases. The mean and third quartile ($\text{Q}_3$) iteration counts are reported above each violin plot.}
    \label{fig:refeed-mechanism-iterations}
\end{figure}

First, Fig.~\ref{fig:refeed-mechanism-iterations} reports the number of iterations required for each Explainer--Verifier pair to produce an accepted explanation. All pairs converged rapidly, with the \texttt{gpt-oss:20b}~$\rightarrow$~\texttt{qwen3:30b} combination performing best, resolving approximately 75\% of cases within one or two iterations. Overall, three of the four pairings successfully verified 63 out of 65 explanations, while \texttt{deepseek-r1:14b}~$\rightarrow$~\texttt{qwen3:30b} verified 64, demonstrating consistently strong correction capabilities across model families. These results highlight the robustness and effectiveness of the refeed mechanism.

From a computational perspective, the cost per explanation can be expressed as $\text{Cost} = 2 + 2K$, where $K$ denotes the number of refinement iterations. As $K$ remains low across cases, each explanation requires approximately 5-7 LLM calls. Assuming an average latency of 30 seconds per call, the total processing time is around 3 minutes per explanation. Therefore, despite the additional verification and refinement steps, the overall computational overhead remains moderate.

\begin{figure}[b!]
    \centering
    \begin{minipage}{1\columnwidth}
        \centering
        \includegraphics[width=\linewidth]{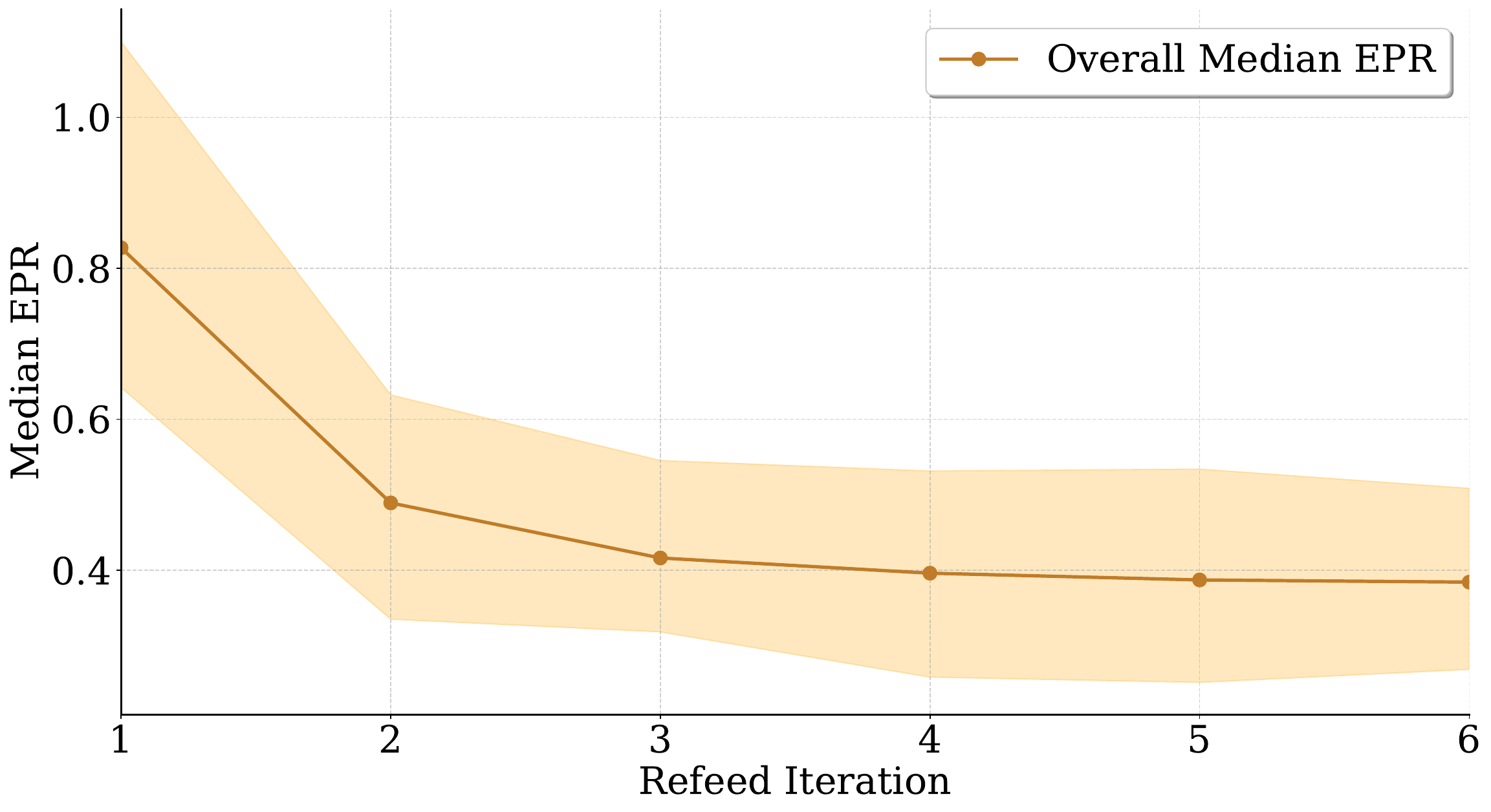}
    \end{minipage}

    \caption{Median Explainer EPR over refinement iterations (1–6) for all ultimately accepted explanations. The shaded bands show the interquartile range.}
    \label{fig:refeed-global-epr}
\end{figure}

To quantify how uncertainty evolves throughout the refinement process (RQ4), we analyzed the Entropy Production Rate. Fig.~\ref{fig:refeed-global-epr} reports the median EPR across all accepted samples. From iterations 1 to 6, the EPR exhibits a monotonic decline under fixed sampling settings (constant temperature and top-$p$), suggesting that Verifier feedback is associated with progressively more stable and coherent Explainer outputs. Because most samples converge within this range, the analysis focused on these early cycles. These findings indicate that the refeed mechanism can contribute to stabilizing the reasoning trajectory while correcting erroneous explanations.

\begin{figure}[t!]
\centering
\includegraphics[width=\columnwidth]{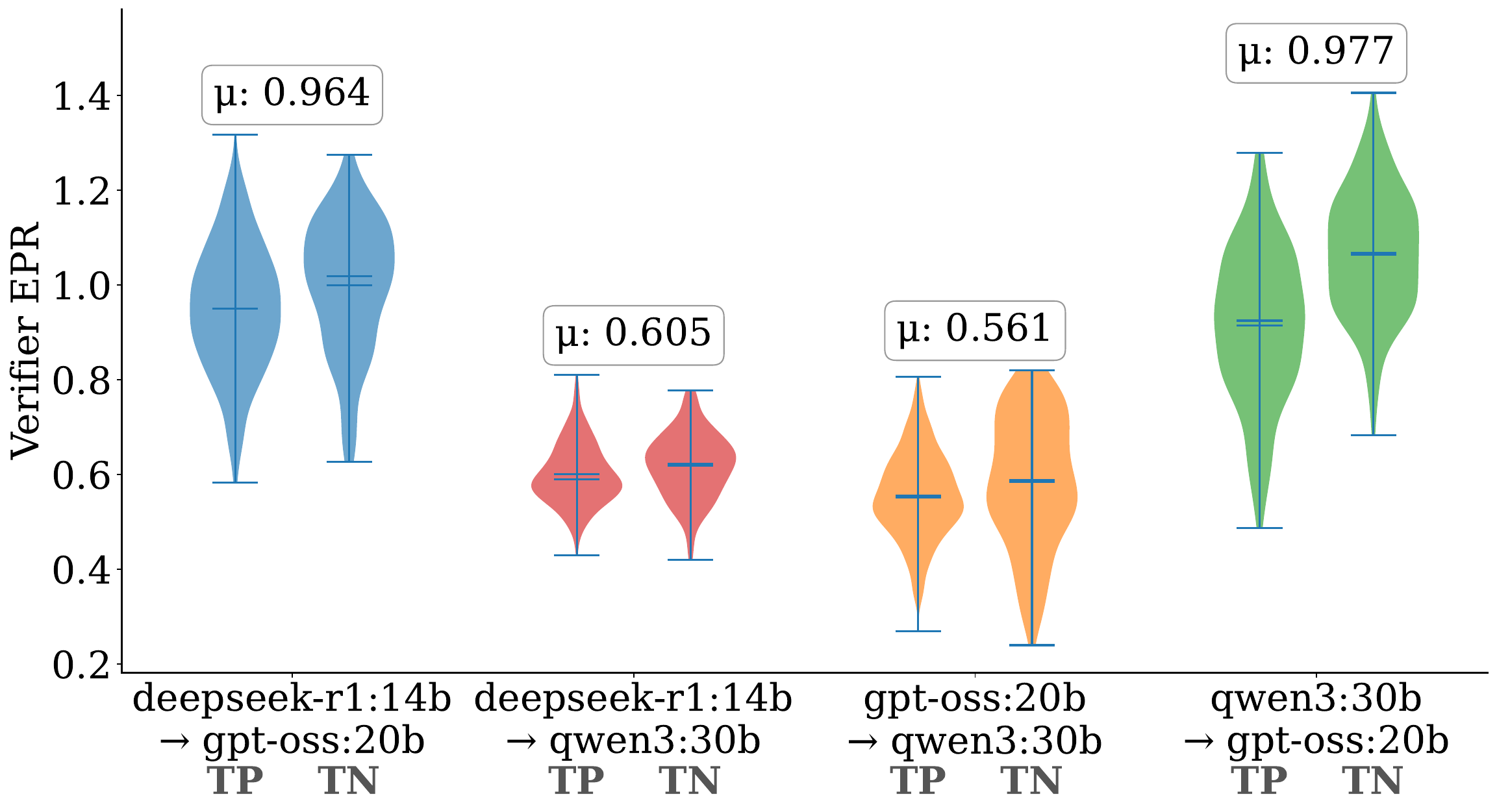}
\caption{Verifier EPR distributions for true positives and true negatives across all Explainer--Verifier pairs.}
\label{fig:verifier_epr}
\end{figure}

A complementary dimension of Verifier performance is the consistency of its decisions, which is also captured by the EPR, where lower values correspond to more stable and predictable verification behavior. As measured in the natural error space and shown in Fig.~\ref{fig:verifier_epr}, \texttt{qwen3:30b} exhibits the lowest and tightest EPR distributions (e.g., $\mu=0.561$, $\sigma=0.102$ for \texttt{gpt-oss:20b}$\rightarrow$\texttt{qwen3:30b}), reflecting reliable judgment patterns. In contrast, \texttt{gpt-oss:20b} operates with substantially higher EPR levels (e.g., $\mu=0.964$, $\sigma=0.138$ for \texttt{deepseek-r1:14b}$\rightarrow$\texttt{gpt-oss:20b}), suggesting a greater internal variability. These results show that \texttt{qwen3:30b} not only achieves the highest verification accuracy but also maintains the most stable decision-making process, underscoring its strength as a Verifier.

\begin{figure}[htbp!]
\centering
\includegraphics[width=\columnwidth]{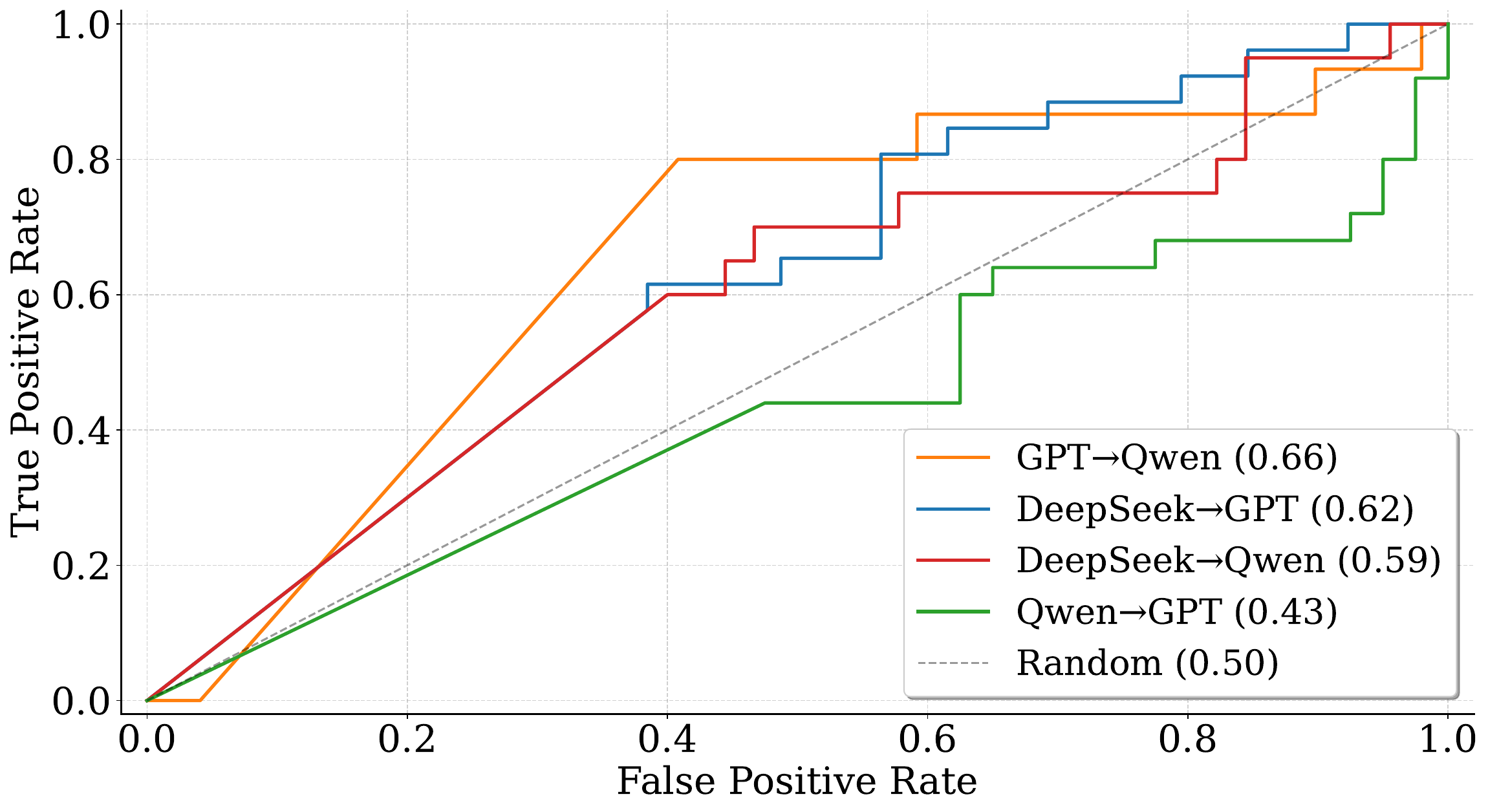}
\caption{ROC curves evaluating how well the initial EPR predicts the need for iterative refinement. Higher AUC values indicate stronger early-warning capability.}
\label{fig:epr_roc}
\end{figure}

Finally, we investigated whether the Explainer’s first-iteration EPR could predict the need for iterative refinement. For this purpose, the initial EPR was treated as a continuous predictor, and “needs correction” was defined as a binary outcome: explanations requiring $\geq$3 refinement cycles or ultimately rejected were labeled 1; those accepted within 1–2 cycles were labeled 0. According to Fig.~\ref{fig:epr_roc} that presents the resulting ROC curves, the \texttt{gpt-oss:20b}~$\rightarrow$~\texttt{qwen3:30b} pair exhibited the strongest predictive power (AUC = 0.66), indicating that a higher initial EPR may correlate with a greater likelihood of requiring correction. The \texttt{deepseek-r1:14b}~$\rightarrow$~\texttt{gpt-oss:20b} (AUC = 0.62) and \texttt{deepseek-r1:14b}~$\rightarrow$~\texttt{qwen3:30b} (AUC = 0.59) pairs provided weaker but still useful signals. In contrast, \texttt{qwen3:30b}~$\rightarrow$~\texttt{gpt-oss:20b} performed below the random baseline (AUC = 0.43). These findings show that the EPR can serve as an early-warning indicator of explanation quality, although its predictive utility is highly pair-dependent.

\section{Conclusions}

This study introduced a Two-Stage LLM Meta-Verification Framework designed to improve both the accessibility and reliability of XAI explanations. The framework consists of three core components: (i) an Explainer LLM that converts raw XAI outputs into natural-language explanations; (ii) a Verifier LLM that inspects these explanations for logical inconsistencies, attribution mismatches, omissions, hallucinations and structural errors; and (iii) an iterative refeed mechanism that automatically revises flawed explanations. Experiments conducted across five XAI methods and datasets, using three LLM families, demonstrated that meta-verification substantially improves explanation faithfulness, increasing accuracy from 59–77.8\% for Explainer-only outputs to 81.8–95.21\% after verification. The framework also enhances linguistic accessibility by simplifying complex, raw XAI outputs. Furthermore, the EPR analysis demonstrated that the Verifier’s feedback progressively guided the Explainer toward more stable and coherent reasoning and that initial EPR scores could help anticipate when refinement was required. Collectively, these findings establish the framework as a practical and effective foundation for developing trustworthy and democratized XAI systems.

Despite its strengths, the proposed framework remains sensitive to prompt design and model-specific biases, motivating further exploration of prompt-engineering strategies, as well as broader evaluations across diverse LLM architectures. Future work may also explore fine-tuning both the Explainer and the Verifier on curated explanation–verification datasets. Moreover, as all XAI outputs are currently converted to text, extending the framework to multimodal foundation models capable of directly interpreting visual XAI artifacts is a promising direction. Finally, while the framework assesses dimensions such as correctness, completeness and linguistic accessibility, dedicated user studies with human participants are needed to evaluate subjective properties, including trust, perceived usefulness and overall satisfaction.

\bibliography{references.bib}
\bibliographystyle{IEEEtran}

\end{document}